\documentclass[10pt,twocolumn,letterpaper]{article}

\usepackage[pagenumbers]{cvpr} 

\definecolor{cvprblue}{rgb}{0.21,0.49,0.74}
\usepackage[pagebackref,breaklinks,colorlinks,allcolors=cvprblue]{hyperref}

\usepackage{hyperref}
\usepackage{booktabs}
\usepackage{multirow}
\usepackage[table]{xcolor}
\usepackage[font=small,labelfont=bf]{caption}  
\usepackage{enumitem}
\usepackage{caption}
\title{PhysiGen: Integrating Collision-Aware Physical Constraints for High-Fidelity Human-Human Interaction Generation}

\author{
\textbf{Nan Lei}\textsuperscript{1},
\quad \textbf{Yuan-Ming Li}\textsuperscript{1},
\quad \textbf{Ling-An Zeng}\textsuperscript{1},
\quad \textbf{Liang Xu}\textsuperscript{3}, \\
\quad \textbf{Zhi-Wei Xia}\textsuperscript{1},
\quad \textbf{Hui-Wen Huang}\textsuperscript{4},
\quad \textbf{Fa-Ting Hong}\textsuperscript{2$\star$},
\quad \textbf{Wei-Shi Zheng}\textsuperscript{1}\\
\textsuperscript{1} \small Sun Yat-sen University \quad
\textsuperscript{2} \small The Hong Kong University of Science and Technology \\
\textsuperscript{3} \small Shanghai Jiao Tong University \quad
\textsuperscript{4} \small Guilin University of Electronic Technology \\
{\tt\small lein7@mail2.sysu.edu.cn; fhongac@connect.ust.hk}
}

\begin{document}
\maketitle

{
  \renewcommand{\thefootnote}{\fnsymbol{footnote}}
  \footnotetext[1]{Corresponding author.}
}

\begin{abstract}

Despite substantial progress in text-driven 3D human motion synthesis, generating realistic multi-person interaction sequences remains challenging. 
Notably, body interpenetration is a pervasive issue from both data acquisition to the generated results, which significantly undermines the realism and usability. Previous generative models either ignored this issue or introduced computationally expensive mesh-level loss functions to alleviate inter-body collisions. 
In this paper, we propose a general-purpose and computationally efficient optimization strategy named PhysiGen to explicitly integrate collision-aware physical constraints for human-human interaction generation. 
Specifically, we simplify the high-resolution human body mesh into geometric primitives to greatly reduce the cost of inter-person collision detection. Moreover, we identify the collision regions as the guidance of the optimization directions. 
PhysiGen is plug-and-play and can be readily integrated into existing human interaction generation models. Extensive cross-dataset and cross-model experiments show that our method can effectively reduce interpenetration and significantly improve visual coherence and physical plausibility compared to the state-of-the-art methods. Code \& Data: \href{https://github.com/iSEE-Laboratory/PhysiGen.git}{https://github.com/iSEE-Laboratory/PhysiGen.git}.
\end{abstract}    

\vspace{-5mm}
\section{Introduction}
Human motion synthesis~\cite{survey} aims to generate natural and coherent 3D human motion sequences based on conditions such as text~\cite{text1,text2,text3,zeng2025progressive,li2025irg,zeng2025light}, audio~\cite{audio2,audio1}, and scenes~\cite{scene1,scene2}, which has broad applications in AR/VR~\cite{ar1,ar2,huang2025modeling}, gaming~\cite{game1,game2}, and robotics~\cite{robot1,robot2,huang2026learning}. Diffusion models propel single-person motion generation to achieve unprecedented precision. Building upon this, more attempts extend to two-person~\cite{interhuman,related:in2in}, or multi-person interaction scenarios~\cite{shan2024towards,multi-person1}, where the challenges lie in modeling inter-person relationships. The intricate interaction patterns and spatiotemporal coherence impose higher demands on sustaining physical realism, reasonable interactions, and avoiding interpenetration.

\begin{figure}[t]
  \centering
  \includegraphics[width=\columnwidth ]{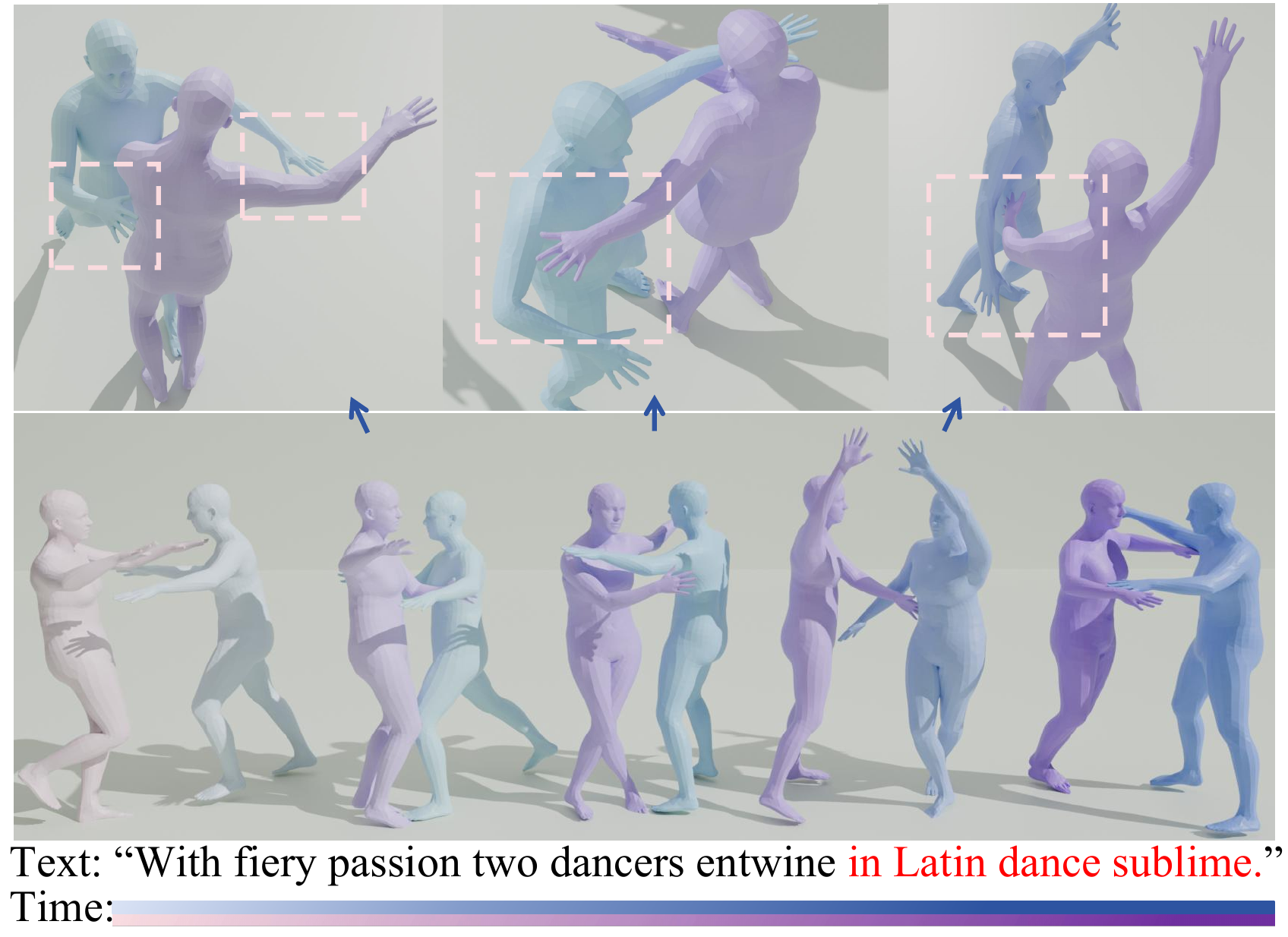}
  \captionof{figure}{Our proposed PhysiGen can generate realistic human interaction motions with minimal interpenetration. Top: close-up views show physically plausible contact without severe collisions. Bottom: full motion sequence demonstrates fluid interaction and semantic consistency with the text.}
  \vspace{-1mm}
  \label{fig:1}
\end{figure}

Current mainstream two-person motion generation approaches ignore the physical constraints of inter-person interactions, which results in body interpenetration and undermines the semantic plausibility of interactions and usability for downstream applications.
Existing methods for human-object/scene interaction synthesis~\cite{related:diffGrasp,related:resolving} adopt mesh-level collision losses based on SDF (Signed Distance Function)~\cite{sdf} to alleviate collisions. However, these mesh-level collision calculations are computationally expensive and time-consuming, especially for human-human interactions, as each human contains thousands of vertices. 
For instance, incorporating the SDF-based loss into the InterGen~\cite{interhuman} model leads to a dramatic training time increase—from 3 days to 14 days, which is practically unacceptable for large-scale deployment.
~\cite{cheng2019occlusion} focuses on extracting 3D human poses from video, which uses cylinders only to generate binary occlusion labels as input(data augmentation technique), without performing collision computation or any motion generation optimization.

\begin{figure*}[!ht]
  \centering
  \includegraphics[width=\textwidth ]{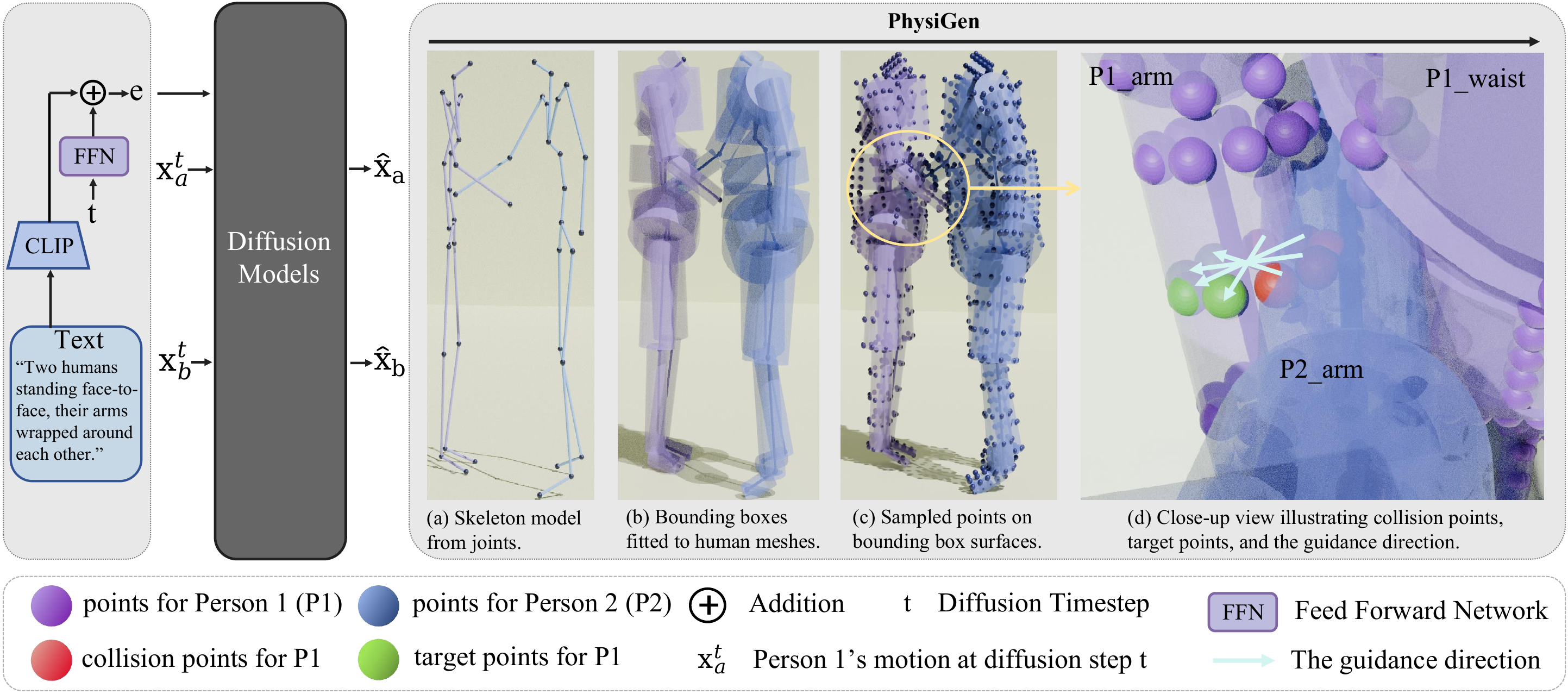}
  \caption{\textbf{Overview of the PhysiGen Framework.} For each detected collision point (red) on Person 1, we compute its corresponding antipodal target point (green) and derive a guidance direction (light blue arrow) to reduce interpenetration. PhysiGen guides the model to adjust the poses along these guidance directions to eliminate interpenetration.}
  \label{fig:2}
  \vspace{-3mm}
\end{figure*}

To address this challenge, we propose a plug-and-play and efficient collision optimization strategy called PhysiGen. 
This method can be easily integrated into existing human motion generation frameworks without modifying the original architecture or adding significant training cost.
First, we model the high-resolution human body mesh as simplified geometric primitives that compactly enclose the human skeletons, such as cylinders or cuboids. 
Compared to directly operating on thousands of mesh vertices, this geometric representation is much simpler and maintains a high-fidelity spatial approximation of human shape, significantly reducing the complexity of collision detection.

And then, we introduce a physics-inspired repulsion mechanism to simulate the physical forces between human body surfaces. 
Specifically, when two individuals come into contact and exhibit signs of interpenetration, PhysiGen computes a clear optimization direction based on the spatial relationship of sampled points on their geometric proxies. This direction effectively acts as a repulsive force, guiding the colliding points to move apart.
By continuously applying this optimization signal during training, PhysiGen encourages the model to learn motion patterns and poses that naturally avoid penetration, ultimately leading to multi-human interactions that are more physically plausible.
Moreover, since this strategy is geometry-based and modular by design, it exhibits strong scalability and generality, making it easy to integrate with various human body structures, interaction scenarios, and even multimodal generative frameworks.

Our design is plug-and-play and can be easily deployed to existing human interaction generation models. Experiments on the large-scale human-human interaction datasets InterHuman and Inter-X demonstrate that our method can effectively reduce interpenetration while improving motion coherence and semantic consistency.

In summary, our main contributions are as follows:
\begin{itemize} [noitemsep, topsep=0pt] 
    \item  
    We develop a plug-and-play optimization strategy called PhysiGen, aimed at efficiently addressing body interpenetration in human-human interaction generation. By simplifying complex human meshes into geometric proxies, our method significantly reduces the computational cost of collision detection(Fig.~\ref{fig:1}).

    \item  The core of PhysiGen is its special direction-guided method. It finds collision points and calculates guidance vectors to give a useful signal for training, which helps the model generate more physically correct motions.

 \item We conduct extensive experiments on the InterHuman and Inter-X datasets with multiple existing state-of-the-art generative models, and the results demonstrate the effectiveness and universality of our PhysiGen.

\end{itemize}

\section{Related Work}

\noindent \textbf{3D Human Motion Generation.}
Early 3D human motion generation models~\cite{xu2023actformer,related:Motionclip,related:Temos,related:T2M,related:AttT2M-54} mainly relied on variational autoencoders (VAEs)~\cite{related:VAE} to learn a shared latent space between text and motion, later enhanced by integrating Transformers to improve expressiveness. Recently, influenced by autoregressive methods and large Transformers in the language domain~\cite{related:Transformers,related:Transformers-10,related:Transformers-53}, motion generation adopted tokenization and autoregressive prediction to improve realism~\cite{related:tm2t-19,related:AttT2M-54}.
In recent years, diffusion models have become mainstream due to their superior generative performance, with methods leveraging probabilistic mapping~\cite{related:Motiondiffuse,related:MDM}, combined with Transformers~\cite{related:Flame,related:Motiondiffuse}, denoising strategies~\cite{related:Motiondiffuse}, and latent diffusion to significantly improve results and efficiency~\cite{related:MLD}.

\noindent \textbf{Human-Human Interaction Generation.}
In recent years, text-driven human interaction motion generation has also attracted attention.
Commdm~\cite{related:ComMDM} combines small-scale two-person data with pre-trained text diffusion models to generate asymmetric interactive motions;
Intergen~\cite{interhuman} released a large-scale two-person dataset with text annotations called InterHuman and proposed a diffusion model with shared weights and multiple regularization losses; Inter-X~\cite{related:inter-x} introduces a larger human-human interaction dataset with versatile tasks~\cite{xu2024regennet}. 
Momat-Mogen~\cite{related:Momat-Mogen} extended retrieval-based diffusion models for interactive motion generation.
More recently, In2in~\cite{related:in2in} and Intermask~\cite{related:Intermask} further improved interactive motion generation. 
TIMotion~\cite{related:TIMotion} enhances performance by injecting causal interactions, employing Role-Evolving Scanning and using Localized Pattern Amplification.
Although these methods have significantly improved generation quality, interpenetration issues in human-human motions remain widespread. 

\noindent \textbf{Penetration Loss of Motion Generation.}
To avoid implausible penetrations when generating human-object/scene interactions, existing works primarily address this issue by introducing contact constraints or geometric loss functions.
Studies such as~\cite{related:CG-HOI, related:Gaze-guided, related:Hoi-diff} focus on dynamically guiding the generation process during inference. They compute the distance between predicted human joints or surface points and the object mesh (e.g., using SDF-based losses) as penalty terms to pull the generated trajectories back to physically plausible contact configurations.
Other works like~\cite{related:diffGrasp, related:resolving, related:HOC} incorporate penetration losses during training, typically defined as the Signed Distance Function (SDF) between key vertices on the human or hand mesh and the object or scene, or alternatively, as the Euclidean distance between human and object vertices. 
However, these approaches often suffer from high computational overhead due to mesh-level operations, making them less practical for human-human motion generation. 
To address this critical challenge, we propose an efficient collision optimization method designed to effectively reduce interpenetration and enhance the physical plausibility of generated motions.

\section{PhysiGen}

\subsection{Problem Definition.}

Given a textual description, our goal is to train a generative model that can synthesize 3D motion sequences ${M_{a}, M_{b}}$ for two interacting persons that conform to the semantics expressed in the text and have a fixed sequence length. Following the data representation used in InterGen~\cite{interhuman}, our representation includes global joint positions $j_g^p \in \mathbb{R}^{3N_j}$, velocities $j_g^v \in \mathbb{R}^{3N_j}$ in the world frame, 6D representation of local rotations $ j^r \in \mathbb{R}^{6N_j} $ in the root frame, and binary foot-ground contact features $ c^f \in \mathbb{R}^4 $, where $N_j$ denotes the joint number.

For Inter-X, we follow the data representation of ~\cite{related:inter-x} with an additional inclusion of global joint positions. Specifically, it includes a 6D representation of local rotations $ j^r \in \mathbb{R}^{6N_j}$ in the root frame and global joint positions $j_g^p \in \mathbb{R}^{3N_j}$, where $N_j$ denotes the joint number.

\subsection{Overview of PhysiGen}
\noindent \textbf{Approximating Mesh with Bounding Box.} To avoid the high computational cost of collision detection in traditional human mesh models, we approximate the complex human body mesh with simplified geometric primitives, such as cylinders and cuboids, whose parameters are pre-optimized to tightly fit individual body shapes, ensuring accurate yet efficient representation.
\noindent \textbf{Collision Points Detection.} Based on these simplified geometric bodies, we have employed an efficient mathematical method to quickly identify collision points between the humans. This approach has significantly reduced computational complexity, enabling fast collision detection.
\noindent \textbf{Guidance Direction Calculation.} For each detected collision point, we construct a guidance vector that defines an explicit escape path from the penetrated region. This vector then directly informs a differentiable guidance loss, providing a corrective gradient that guides the model optimization process to mitigate physical penetration.

\subsection{Fitting Bounding Boxes to the Human Mesh}
\label{sec:3.3}
\begin{figure}[t]
  \centering
  \includegraphics[width=\columnwidth ]{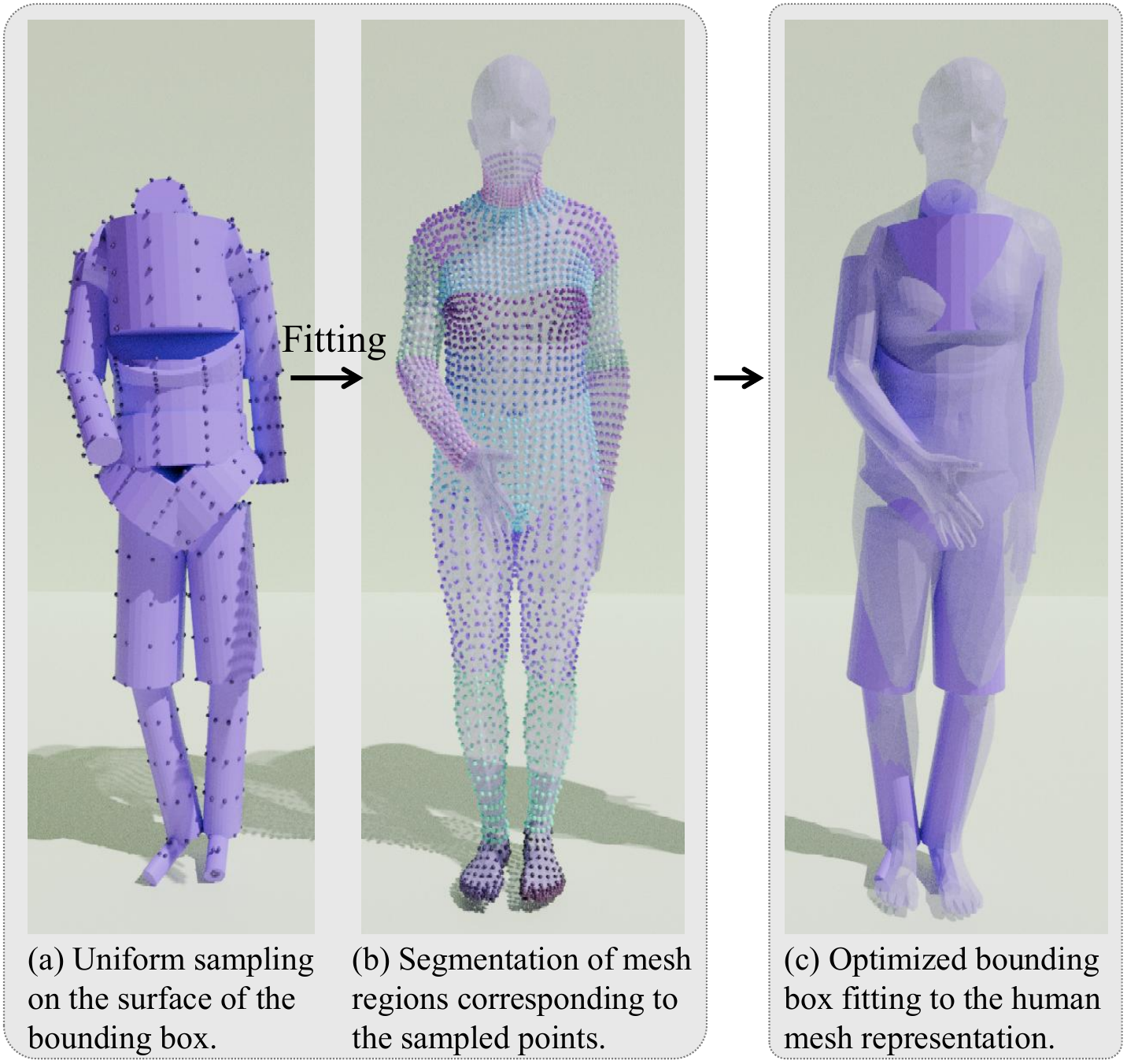}
  \caption{Overview of the bounding box fitting process and the final volumetric proxy representation.}
  \label{fig:3}
\end{figure}

To make collision-aware optimization fast and accurate, PhysiGen first simplifies the complex human body shape using bounding boxes.
This section explains how we build and adjust these boxes to match different human body sizes and shapes.
Specifically, we construct a skeletal model based on the topological connections of human joints (as shown in Fig.\ref{fig:2}(a)). 
Then, around each skeletal segment, we generate bounding boxes composed of cuboids or cylinders, which aim to roughly fit the contour of the real human mesh (as shown in Fig.\ref{fig:3}(a)). 
It is worth noting that the orientation of the cuboids is determined by the two hip joints and one spine joint, whereas cylinders have no such restriction.
Subsequently, we uniformly sample points on the surface of each bounding box (Fig.\ref{fig:3}(a)),
and segment the points on the real human mesh into regions, so that each mesh region corresponds to a sampled point on the bounding box.
As illustrated in Fig.\ref{fig:3}(b), different regions are marked using different colors to clearly show this segmentation.
Next, we calculate the minimum distance between each pair of “bounding box sampled point and corresponding mesh region points,” summing these distances as loss, which is defined as:
\begin{equation}
\mathcal{L}_{\text{fit}} = \sum_{j=1}^{M} \sum_{q \in \mathcal{S}_j} \min_{p \in \mathcal{P}_j} \|q - p\|^2,
\end{equation}
where $\mathcal{L}_{\text{fit}}$ denotes the fitting loss, $M$ is the number of bounding boxes, $\mathcal{S}_j$ is the set of sampled surface points on the $j$-th box, and $\mathcal{P}_j$ is the set of mesh points corresponding to the region covered by the \(j\)-th bounding box; The loss measures the squared Euclidean distance between each sampled point $q \in \mathcal{S}_j$ and its nearest mesh point $p \in \mathcal{P}_j$, encouraging geometric proxies to fit the human mesh closely.

We iteratively update the size of the bounding boxes using optimization algorithms (e.g., gradient descent) to minimize \( \mathcal{L}_{\text{fit}} \). The optimization stops when the loss converges and the bounding box parameters stabilize, and then, we obtain bounding box parameters that best fit the human body volume (Fig.\ref{fig:3}(c)). 
Considering individual differences, we learn a separate set of bounding box parameters for each body shape, ensuring the model’s good generalization and accuracy across diverse human body sizes and shapes.

\subsection{Methodology Details of PhysiGen.}

\noindent \textbf{Approximating Mesh with Bounding Box.}
To achieve efficient collision detection, we first simplify the complex human body by using simplified geometric proxies. 
Following the same approach described in Sec.~\ref {sec:3.3}, we construct a skeletal model based on the topological connections between human joints (as shown in Fig.\ref{fig:2}(a)). 
Next, we generate bounding boxes around each skeletal segment to approximate local body regions (as shown in Fig.\ref{fig:2}(b)). 
We then uniformly sample points on the surface of each bounding box to create a set of candidate points (as shown in Fig.\ref{fig:2}(c)). 
These sampled points represent the spatial distribution of the human body surface and maintain a relative relationship with the skeleton’s position for real-time collision tracking.

\noindent \textbf{Collision Points Detection.}
After fitting bounding boxes to the human mesh, we can perform collision detection efficiently.
Specially, for each sampled point $p_i$ on Person 1’s geometry, we will check whether it falls inside any bounding box of Person 2.
If it does, we consider it a collision and mark $p_i$ as a collision point of Person 1.

The collision checking rules are simple:
A point \( p \) is considered to be inside a cuboid if it lies between all three pairs of opposite faces.
For each face pair (along the width, height, and depth directions), we compute two vectors \( \mathbf{v}_1 \) and \( \mathbf{v}_2 \) from the point \( p \) to the vertices (or representative points) of the two opposite faces.  
If the dot product of these two vectors is negative, it indicates that point \( p \) lies between that pair of faces.
If this condition is satisfied for all three face pairs, then the point \( p \) is inside the cuboid, which can be expressed:
\vspace{-3.5pt}
\begin{equation}
\setlength{\belowdisplayskip}{8pt}
(p - v_1^k) \cdot (p - v_2^k) < 0, \quad \text{for } k \in \{x, y, z\} ,
\end{equation}
where \( v_1^k \) and \( v_2^k \) are diagonal vertices on opposite faces along axis \( k \).

A point \( p \) is considered to be inside a cylinder if it satisfies both of the following conditions: the distance from \( p \) to the cylinder axis is less than the radius \( r \), and the projection of \( p \) onto the axis lies between the bottom and top faces (i.e., within the height range), which can be expressed as:
\vspace{-3.5pt}
\begin{equation}
\setlength{\belowdisplayskip}{8pt}
\text{dist}(p, \text{axis}) < r
\quad \text{and} \quad
0 \leq (p - a) \cdot \hat{\mathbf{v}} \leq h ,
\end{equation}\
where \( a \) is the bottom center of the cylinder, \( \hat{\mathbf{v}} \) is the unit axis direction, \( h \) is the height, and \( \text{dist}(p, \text{axis}) \) is the perpendicular distance from \( p \) to the axis.
The two methods described above allow fast, batch-style collision detection between simplified geometric proxies of two people.

\begin{figure}[t]
  \centering
  \includegraphics[width=\columnwidth ]{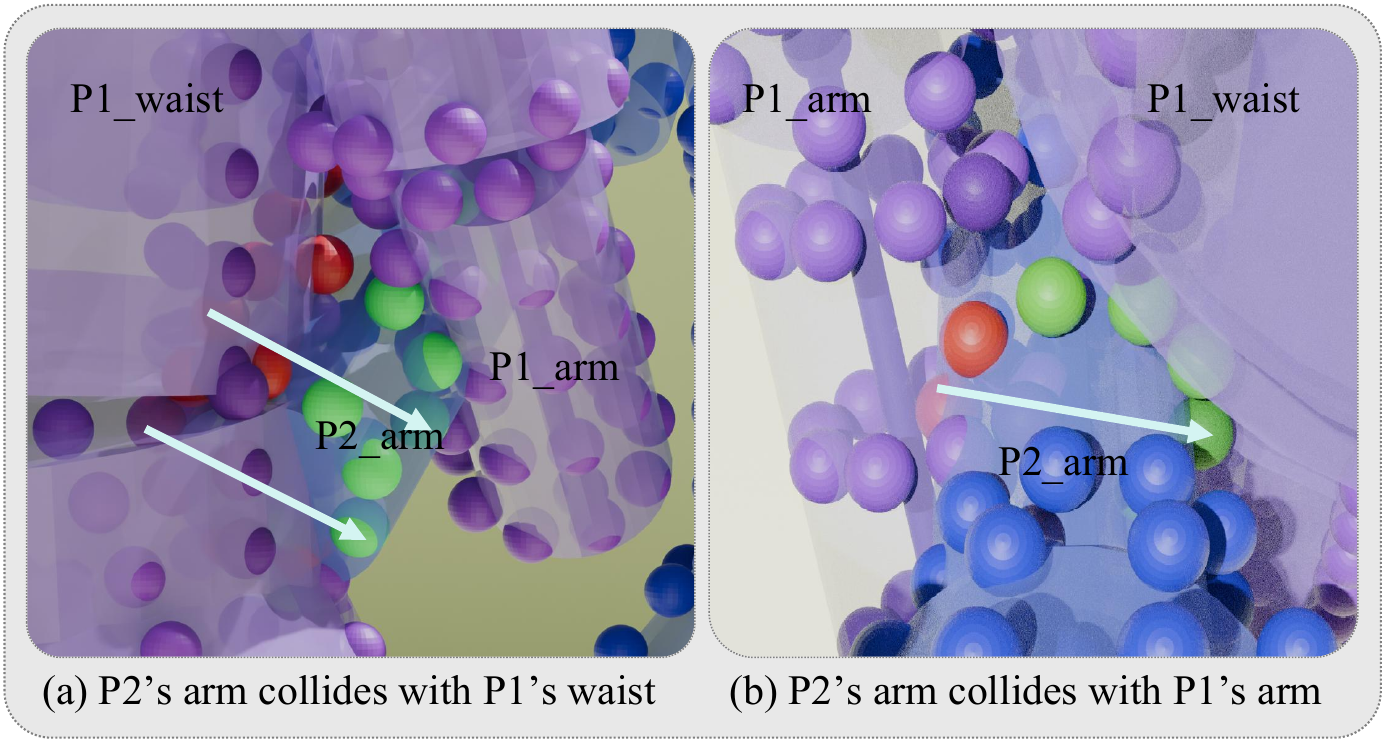}
  \caption{\textbf{Illustration of Multi-Region Simultaneous Collision Handling.} The figure demonstrates the detected collision points (red), corresponding target points (green), and guidance directions (arrows) guiding the model to avoid interpenetration, specifically when Person 2's (P2) arm simultaneously penetrates Person 1's (P1) waist (left) and arm (right). This showcases PhysiGen's effective capability in handling complex multi-region collisions.}
  \label{fig:4}
\end{figure}

\noindent \textbf{Guidance Direction Computation.}
Once the set of collision points is detected, our PhysiGen framework will compute a guidance direction for each point to resolve the penetration. 
Specifically, the guidance direction is a \textit{guidance vector} for each collision point, which is designed to represent an ``escape path", a direction of steepest descent away from the collision state.
Directly calculating the ``escape path" is challenging. 
Therefore, we propose an extremely efficient analytical heuristic to approximate this gradient: By leveraging the intrinsic symmetry of the proxy geometry, we find the antipodal point of the collision point on its geometry and construct a stable, noise-free vector:  
\begin{equation}
\setlength{\belowdisplayskip}{8pt}
\mathbf{d_i} \;=\; \frac{q_i-p_i}{\|\, q_i-p_i \,\|} ,
\end{equation}
where \( p_i \) is the detected collision point and \( q_i \) is the corresponding target point on the opposite surface of the colliding geometric proxy.
The vector \( \mathbf{d}_i \) is then used as the optimization direction for the collision loss \( \mathcal{L}_{coll} \):
\begin{equation}
\vspace{-0.5mm}
\setlength{\belowdisplayskip}{8pt}
\mathcal{L}_{coll} = \sum_{i:\,p_i\in\text{collision}}
\bigl\|\,sg[(p_i + \mathbf{d}_i)] - p_i\bigr\|^{2}_{2},
\end{equation}
where $sg[\cdot]$ denotes the stop-gradient operation. As shown in Fig.~\ref{fig:2} (d), Person 1’s arm penetrates Person 2’s chest during the motion. In the figure, the red dot marks the detected collision point on $P_1$, the green dot marks its corresponding target point $q_i$, and the arrow indicates the unit vector $d_i$. 
By minimizing $L_{coll}$, the network automatically fine-tunes $P_1$’s arm along the arrow direction, thereby effectively eliminating the unreasonable interpenetration while preserving the hugging semantics and motion continuity.

\noindent \textbf{Multi-Region Collision Avoidance.} When one body part of a person simultaneously collides with multiple parts of another person, the guidance vectors generated at different contact points may diverge or even conflict.
Taking the scenario illustrated in Fig.\ref{fig:2}(d) as an example, if Person 2's arm penetrates both Person 1's waist Fig.\ref{fig:4}(a) and Person 1's arm Fig.\ref{fig:4}(b), both areas are flagged as collision zones; these collision points may individually suggest opposing optimization directions. 
PhysiGen addresses this by aggregating these local vectors to determine a unified guidance direction.
Since the spatial density of these points inherently reflects the severity of the interpenetration, the aggregated vector $\mathbf{d}_{\text{total}}$ naturally biases away from the most heavily congested areas.
In Fig.\ref{fig:4}, because the waist has more collision points, its corresponding collision vector is given a higher weight, causing the final synthesized avoidance direction to be biased away from Person 1's waist area.
Specifically, if there are \( n_w \) collision points on the waist and \( n_a \) collision points on the arm, the total optimization direction \(\mathbf{d}_{\text{total}}\) can be expressed as:
\begin{equation}
    \mathbf{d}_{\text{total}} = \frac{n_w \mathbf{d}_w + n_a \mathbf{d}_a}{n_w + n_a},
\end{equation}
where \(\mathbf{d}_w\) and \(\mathbf{d}_a\) are the unit vectors from the waist and arm collision points toward their respective target points.
As shown in Fig.\ref{fig:2}(d), Person 1's arm is also slightly adjusted toward this combined avoidance direction. 
Under this optimization, the arm of Person 2 moves away from Person 1's waist and arm regions, while Person 1's arm moves away from Person 2's arm region. 
This process rapidly converges and effectively eliminates multiple interpenetrations while preserving the semantics of hugging or interaction.

\subsection{Loss Integration and Training Strategy}

Our PhysiGen method is that it can be easily added to the training process of existing motion generation models to efficiently reduce interpenetration issues in human-human interactions. Specifically, we consider two training modes:

\begin{itemize}
\item \textbf{From scratch}: PhysiGen's collision loss term directly integrates into the model's entire training process. This enables the model to internalize physical constraints from the initial stages, thereby naturally avoiding collisions. 

\item \textbf{Adaption}: PhysiGen acts as a "plug-in" optimization solution for models already pre-trained on large datasets. By adding our collision loss term during fine-tuning, the model enhances physical realism and reduces interpenetration artifacts while retaining its original semantic understanding and generation capabilities. 
\end{itemize}

\noindent Our PhysiGen does not rely on a specific base model. By combining $\mathcal{L}_{\text{coll}}$ and the original loss $\mathcal{L}_{\text{general}}$ in the base model, the overall training loss can be written as:

\vspace{-3.5pt}
\begin{equation}
\setlength{\belowdisplayskip}{8pt}
    \mathcal{L}_{\text{total}} = \mathcal{L}_{\text{general}} + \lambda_{\text{coll}} \mathcal{L}_{\text{coll}},
\end{equation}
where $\lambda_{\text{coll}}$ is a hyperparameter that balances the two terms.

\begin{table*}[ht]
  \centering
  \resizebox{\textwidth}{!}{
    \begin{tabular}{lccccccccc}
    \toprule
    \multicolumn{1}{l}{\multirow{2}{*}[-0.8ex]{\textbf{Methods}}} &
    \multicolumn{1}{c}{\multirow{1}{*}[-0.8ex]{$coll_{dis}\!\downarrow$}} &
    \multicolumn{1}{c}{\multirow{1}{*}[-0.8ex]{$coll_{ro}\!\downarrow$}}  & 
    \multicolumn{3}{c}{R Precision $\uparrow$}   &
    \multicolumn{1}{c}{\multirow{2}{*}[-0.8ex]{FID $\downarrow$}}      &
    \multicolumn{1}{c}{\multirow{2}{*}[-0.8ex]{MM Dist $\downarrow$}}  & 
    \multicolumn{1}{c}{\multirow{2}{*}[-0.8ex]{Diversity $\to$}}      &
    \multicolumn{1}{c}{\multirow{2}{*}[-0.8ex]{MModality $\downarrow$}} \\
    \cmidrule(lr){4-6}
          & m & \% & Top 1 & Top 2 & Top 3 & & & & \\
    \midrule
    \rowcolor{gray!30} \multicolumn{10}{l}{\textit{On the InterHuman dataset.}}\\
    \midrule
    Real  & $1.134$ & $0.1829$ & $0.452^{\pm.008}$ & $0.610^{\pm.009}$ & $0.701^{\pm.008}$ & $0.273^{\pm.007}$ & $3.755^{\pm.008}$ & $7.948^{\pm.064}$ & -- \\
    \midrule
    InterGen~\cite{interhuman}  & 3.905 & 0.2270 & $0.371^{\pm.010}$ & $0.515^{\pm.012}$ & $0.624^{\pm.010}$ & $\textbf{5.918}^{\pm.079}$ & $5.108^{\pm.014}$ & $7.387^{\pm.029}$ & $2.141^{\pm.063}$ \\
      \textbf{+PhysiGen(from scratch) } & $\textbf{1.836}$ & 0.1878 & $\textbf{0.485}^{\pm.006}$ & $\textbf{0.639}^{\pm.006}$ & $\textbf{0.717}^{\pm.004}$ & $6.044^{\pm.086}$  & $\textbf{3.777}^{\pm.001}$ & $\textbf{7.913}^{\pm.033}$  & $\textbf{1.035}^{\pm.026}$\\
    \textbf{+PhysiGen(adaption) }  & 2.433 & $\textbf{0.1672}$ &$0.435^{\pm.004}$ & $0.591^{\pm.005}$ & $0.673^{\pm.005}$ & $6.026^{\pm.097}$ & $3.797^{\pm.001}$ & $7.839^{\pm.025}$ & $1.185^{\pm.036}$ \\
    \midrule
    in2IN~\cite{related:in2in} &3.142& 0.1863 & $0.455^{\pm.004}$ & $0.611^{\pm.005}$ &$0.687^{\pm.005}$ & $\textbf{5.177}^{\pm.103}$ & $3.790^{\pm.002}$ & $7.940^{\pm.030}$ & $1.061^{\pm.038}$ \\
    \textbf{+PhysiGen(from scratch)}  & 2.870 &0.1678      & $0.462^{\pm.005}$ & $0.622^{\pm.005}$ & $0.703^{\pm.005}$ & $5.264^{\pm.062}$ & $3.787^{\pm.001}$ & $7.951^{\pm.030}$ & $1.076^{\pm.032}$ \\
    \textbf{+PhysiGen(adaption)} & \textbf{2.005} & \textbf{0.1503} & $\textbf{0.481}^{\pm.004}$ & $\textbf{0.637}^{\pm.004}$ & $\textbf{0.712}^{\pm.004}$ & $5.269^{\pm.049}$ & $\textbf{3.780}^{\pm.001}$ & $\textbf{7.979}^{\pm.021}$ & $\textbf{1.015}^{\pm.043}$ \\
    \midrule
    \rowcolor{gray!30} \multicolumn{10}{l}{\textit{On the Inter-X dataset.}}\\
    \midrule
    Real  & $3.352$ & $0.3375$ & $0.429^{\pm.004}$ & $0.626^{\pm.003}$ & $0.736^{\pm.003}$ & $0.002^{\pm.0002}$ & $3.536^{\pm.013}$ & $9.734^{\pm.078}$ & -- \\
    \midrule
    InterGen336* & 12.840 & 0.340 & $0.378^{\pm.0030}$ & $0.566^{\pm.0045}$ & $0.676^{\pm.0038}$ & $0.312^{\pm.0193}$ & $3.950^{\pm.0178}$ & $9.040^{\pm.0898}$ & $\textbf{2.806}^{\pm.1105}$ \\
    InterGen468   & 13.479    &0.363  & $0.391^{\pm.0058}$ & $0.577^{\pm.0048}$ & $\textbf{0.687}^{\pm.0038}$ & $\textbf{0.320}^{\pm.0114}$ & $3.859^{\pm.0143}$ & $9.092^{\pm.0618}$ & $2.999^{\pm.0931}$ \\
    \textbf{+PhysiGen(from scratch) } & 11.729  &0.349  & $\textbf{0.392}^{\pm.0042}$  & $\textbf{0.577}^{\pm.0037}$ & $0.684^{\pm.0033}$ & $0.382^{\pm.0210}$ & $3.848^{\pm.0138}$ & $\textbf{9.120}^{\pm.0506}$ & $3.038^{\pm.0921}$ \\
    \textbf{+PhysiGen(adaption)}   & \textbf{10.764}  & \textbf{0.328} & $0.388^{\pm.0001}$ & $0.574^{\pm.0001}$ & $0.685^{\pm.0002}$ & $0.374^{\pm.0013}$ & $3.067^{\pm.0389}$ & $9.079^{\pm.0236}$ & $3.854^{\pm.0026}$ \\
    \midrule
    TIMotion336*~\cite{related:TIMotion} & 14.229 & 0.354 & $0.398^{\pm.0034}$ & $0.591^{\pm.0046}$ & $0.702^{\pm.0054}$ & $0.557^{\pm.0150}$ & $3.756^{\pm.0141}$ & $8.967^{\pm.0630}$ & $\textbf{2.356}^{\pm.0604}$ \\
    TIMotion468 &11.878 &0.352 & $0.413^{\pm.0049}$ & $\textbf{0.605}^{\pm.0051}$ & $0.713^{\pm.0035}$ & $\textbf{0.395}^{\pm.0150}$ & $3.683^{\pm.0184}$ & $9.133^{\pm.0732}$ & $2.413^{\pm.0673}$  \\
    \textbf{+PhysiGen(from scratch)} & \textbf{10.281} & \textbf{0.339}    & $0.410^{\pm.0041}$  & $0.602^{\pm.0037}$ & $0.713^{\pm.0034}$ & $0.453^{\pm.0131}$ & $3.687^{\pm.0184}$ & $9.042^{\pm.0637}$ & $2.488^{\pm.0783}$ \\
    \textbf{+PhysiGen(adaption)} & 10.860    & 0.342    & $\textbf{0.413}^{\pm.0046}$ & $0.603^{\pm.0041}$ & $\textbf{0.714}^{\pm.0036}$ & $0.401^{\pm.0101}$ & $\textbf{3.678}^{\pm.0139}$ & $\textbf{9.076}^{\pm.0812}$ & $2.389^{\pm.0694}$ \\
    \bottomrule
    \end{tabular}
  }
  \caption{\textbf{Quantitative evaluation on the InterHuman\cite{interhuman} and Inter-X\cite{related:inter-x} test set}. Following previous works, we run all evaluations 20 times, except collision and MModality run 5 times. $\pm$ indicates the 95\% confidence interval. Bold indicates the best result. Since some works do not provide the pretrained models or complete code, we reproduced the results of InterGen336* and TIMotion336* (using Transformer modules) based on the unorganized training code and the open-source validation scripts released by the authors of Inter-X~\cite{related:inter-x}.}
  \label{tab:tableA1}
  \vspace{-3mm}
\end{table*}

\section{Experiment}

\subsection{Datasets}

We evaluated PhysiGen on two human-human interaction datasets for text-conditioned interaction generation:
(1) \textbf{InterHuamn}~\cite{interhuman} contains 7,779 two-person interaction sequences and the corresponding text descriptions.
Following AMASS skeletal framework~\cite{AMASS}, motions in InterHuman are represented with 22 joints (including the root).
(2) \textbf{Inter-X}~\cite{related:inter-x} is the largest collection of human interaction data, with 11,388 interaction sequences. The dataset uses 
the SMPL-X~\cite{SMPL-X} 3D human model, which has 54 keypoints to represent body movement, hand gestures, facial expressions, and overall motion. 
Previous methods adopt different data representations for the two datasets, lacking skeletal data required by PhysiGen on the Inter-X dataset. 
Additional processing is applied as detailed in Sec. \ref{sec:4.3}.


\subsection{Evaluation Metrics}

\subsubsection{Physical Plausibility Metrics.}
To evaluate the inter-penetration, we introduce two metrics based on mesh-level Signed Distance Function (SDF) calculations. 
We define the Collision Distance ($coll_{dis}$) as the average penetration depth between the two participants across all frames. 
Furthermore, we calculate the Collision Rate ($coll_{ro}$), which is the proportion of frames within a sequence that exhibit collision. 
Since the InterHuman dataset uses a skeleton structure with 22 joints and does not include hand modeling, the collision evaluation metrics are calculated without considering the hand mesh.
\subsubsection{Generation Quality Metrics.}
Following established practices in motion generation~\cite{interhuman}, we adopt a standard suite of metrics to evaluate the overall quality. (1) Frechet Inception Distance (FID): measures the distributional similarity between generated and real motion features. (2) RPrecision: the semantic consistency between the motions and their corresponding text prompts. (3) Diversity: measures motion diversity in the generated motion dataset. (4) Multimodality (MModality): gauges diversity within the same text. (5) Multi-modal distance (MM Dist): measures the distance between motions and text features.

\begin{figure*}[!ht]
  \centering
  \includegraphics[width=\textwidth ]{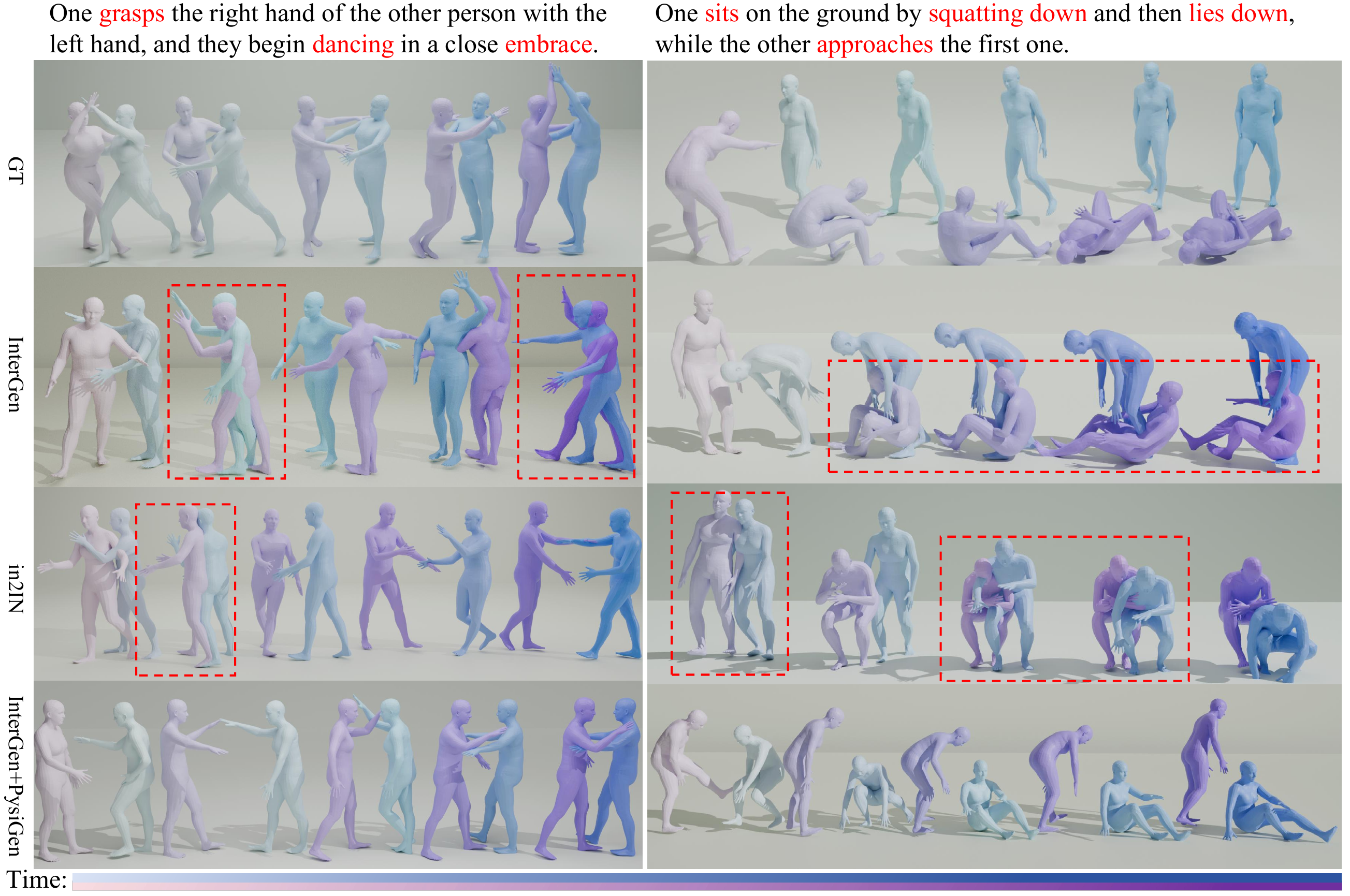}
\caption{\textbf{Qualitative comparison with state-of-the-art methods.} 
The baseline methods suffer from severe artifacts, including body interpenetration ( highlighted in red ) and unnatural or semantically incorrect motions. In contrast, our method generates high-fidelity, physically plausible interactions free of collisions and semantically coherent with the textual descriptions.
}
  \label{fig:5}
  \vspace{-3mm}
\end{figure*}

\subsection{Implementation Details}
\label{sec:4.3}
To apply our strategy to the Inter-X dataset, we made adjustments to the original feature representation. 
First, we removed invalid padding components.
Then, we added the global position of 51 joints, including both hands and all body joints except the root (since the original features already included the root joint), expanding the feature dimension from 336 to 468. 
During testing, we discard the extra dimensions and pad the features back to 336 dimensions to match the original model’s input format, and use the same testing model for evaluation.
Please refer to the supplementary material for more details.


\subsection{Quantitative Analysis} 
We compare the three latest baseline models (InterGen~\cite{interhuman}, in2IN~\cite{related:in2in}, and TIMotion~\cite{related:TIMotion}) and evaluate how PhysiGen enhances the performance in Table~\ref{tab:tableA1}. The relations between the adaptation strategy and train-from-scratch are already discussed in Sec 3.5.

Across both InterHuman and Inter-X datasets, PhysiGen consistently improves physical plausibility and generation quality. For example, on InterHuman, it reduces the collision distance from 390.4 cm to 183.6 cm on InterGen and improves Top-1 R Precision from 0.371 to 0.485. On Inter-X, TIMotion+PhysiGen (from scratch) lowers the collision distance by over 400 cm while raising the Top-1 R Precision from 0.398 to 0.440. These results demonstrate PhysiGen’s strong generalizability and effectiveness with minimal additional cost.

\subsection{Qualitative Analysis} 

Fig.~\ref{fig:5} shows the results of different methods. 
Our PhysiGen results remain fully aligned with the text prompt while significantly reducing interpenetration.
The characters maintain natural contact during the hug and move smoothly during the squat-and-approach sequence, producing realistic, collision-free interactions throughout. 
In short, PhysiGen preserves semantic fidelity and significantly improves physical plausibility compared with prior methods.

\subsection{Ablation Study and Analysis}
In this section, we conduct comprehensive ablation studies to investigate the impact of the collision loss, the effectiveness of the fine-tuning phase, and discuss the effect of the number of surface sampling points on the final results. All ablations are performed on the InterHuman dataset.

\textbf{Geometric Shape Selection.}
Table~\ref{tab:ablation} demonstrates that using cylinders outperforms cuboids in both physical collision metrics and generation quality. This suggests that cylinders better conform to the human mesh, offering more stable and effective optimization guidance. In contrast, cuboids require consideration of the direction of faces, which may result in poor alignment with the body in certain poses, leading to inaccurate optimization directions. In addition, as the number of sampling points increases, collision metrics steadily improve, indicating that denser sampling enables more precise collision detection and provides more guidance vectors to help the model avoid penetrations.

\textbf{Sampling Point Density.}
However, it is worth noting that when the number of sampled points increases from 30 to 50 in cylinders, the generation metrics slightly decline. We believe this is because the collision loss becomes dominant under high-density sampling, causing the model to overly focus on physical constraints and sacrifice part of the motion’s semantic naturalness. Therefore, choosing an appropriate number of sampling points is critical for balancing physical plausibility and generation quality.

\begin{table}[t]
  \centering
\resizebox{\columnwidth}{!}{
\setlength{\tabcolsep}{1mm}
    \begin{tabular}{lccccccc}
    \toprule
             \multicolumn{1}{l}{\multirow{2}{*}[-0.8ex]{{Points\_num}}}
           & $coll_{dis}\!\downarrow$ & $coll_{ro}\!\downarrow$
            & \multicolumn{3}{c}{R Precision $\uparrow$}
            & \multicolumn{1}{c}{\multirow{2}{*}[-0.8ex]{FID $\downarrow$}}  \\
          \cmidrule(lr){4-6}
                & m & \% & Top 1 & Top 2 & Top 3 & &  \\
          \midrule
          Real    & $1.238$ & $0.149$ & $0.452^{\pm.008}$ & $0.610^{\pm.009}$ & $0.701^{\pm.008}$ & $0.273^{\pm.007}$ \\
          \midrule
          in2IN &3.142&	0.186 & $0.455^{\pm.004}$ & $0.611^{\pm.005}$ &$0.687^{\pm.005}$ & 
          $5.177^{\pm.103}$ \\
          \midrule
           \multicolumn{7}{l}{in2IN+PhysiGen (adaption)}\\ 
            \midrule
          
          cuboid\_16 & 2.834   & 0.179    & $\textbf{0.487}^{\pm .0058}$ & $0.637^{\pm .0042}$ & $0.714^{ \pm .0043}$ & $5.135^{ \pm .0700}$       \\
          cuboid\_36  & 2.751    & 0.170    & $0.485^{\pm.0059}$ & $0.636^{\pm.0045}$ & $0.714^{\pm.0044}$ & $\textbf{5.012}^{\pm.0799}$ \\
          cuboid\_64  &2.288  & 0.163      & $0.476^{\pm .0063}$ & $0.631^{ \pm.0052}$ & $0.712^{ \pm .0045}$ & $5.161 ^{\pm .0919}$   \\
           \midrule
          cylinder\_10 & 2.634 & 0.171     & $0.479^{\pm.0054}$ & $0.632^{\pm.0048}$ & $0.712^{\pm.0049}$ & $5.194^{\pm.0597}$ \\
          cylinder\_30  & 2.515   & 0.170& $0.485^{\pm.0047}$ & $\textbf{0.639}^{\pm.0037}$ & $\textbf{0.716}^{\pm.0048}$ & $5.181^{\pm.0688}$ \\
          cylinder\_50  & \textbf{2.005}   & \textbf{0.150}  & $0.481^{\pm.0037}$ & $0.637^{\pm.0043}$ & $0.712^{\pm.0035}$ & $5.269^{\pm.0489}$ \\
          \bottomrule
    \end{tabular}%
    }
    \caption{\textbf{Ablation on the Interhuman test set}. Points\_num indicates the number of sampled points on each cylinder or cuboid. We run all the evaluations 20 times, except collision and MModality run 5 times. $\pm$ indicates the 95\% confidence interval. Bold indicates the best result. All the models employ the same non-canonical representation. }
  \label{tab:ablation}
\end{table}
  
\begin{table}[t]
\centering
\resizebox{\columnwidth}{!}{
\begin{tabular}{lcccccccc}
\toprule
\multirow{2}{*}{\textbf{\shortstack{Setup \\ num\_points}}} & \multirow{2}{*}{w/o loss}
& \multicolumn{3}{c}{\textbf{PhysiGen}} 
& \multirow{2}{*}{\shortstack{w/o loss \\ + root}}
& \multicolumn{3}{c}{SDF loss} \\
\cmidrule(lr){3-5} \cmidrule(lr){7-9}
 & & \textbf{10$\times$19} & \textbf{30$\times$19} & \textbf{50$\times$19} 
 & & 128 & 1024 & 6890 \\
\midrule
\textbf{Memory (MB)} & 15107 & \textbf{16045} & \textbf{18093} & \textbf{20219} & 15172 & 24152 & 24154 & 24156 \\
\textbf{Time (s)} & - & \textbf{0.013} & \textbf{0.033} & \textbf{0.053} & - & 0.352 & 0.685 & 3.734 \\
\bottomrule
\end{tabular}
}
\caption{\textbf{GPU memory and training time for a single batch under different loss configurations.} All tests were performed on a single NVIDIA L40 GPU with a batch size of 16. The training time is an average over 20 batches. 
}
\label{tab:mem_time_ablation}
\end{table}

\subsection{Comparison of Computational Costs}

To comprehensively evaluate the computational overhead of our proposed method (PhysiGen) compared to SDF loss, we measured both GPU memory consumption and per-batch training time on the InterGen model with the InterHuman dataset. Since computing the SDF loss requires the root joint’s 6D rotation to obtain the human mesh, and the original InterGen input does not include this information, we added this dimension for training. As shown in the Table~\ref{tab:mem_time_ablation} “+ root", we report the GPU memory usage after adding this extra dimension. 

Our findings show that incorporating PhysiGen introduces only a marginal increase in memory usage and training time. For instance, using 50×19 sampling points, PhysiGen increases memory usage by about 5GB compared to the baseline, yet the training time per batch remains low at just 0.053 seconds per batch. This suggests that PhysiGen scales efficiently even as the number of sampled points increases, and is well-suited for large-scale training.

\begin{figure}[htbp]
    \centering
    \includegraphics[width=\linewidth]{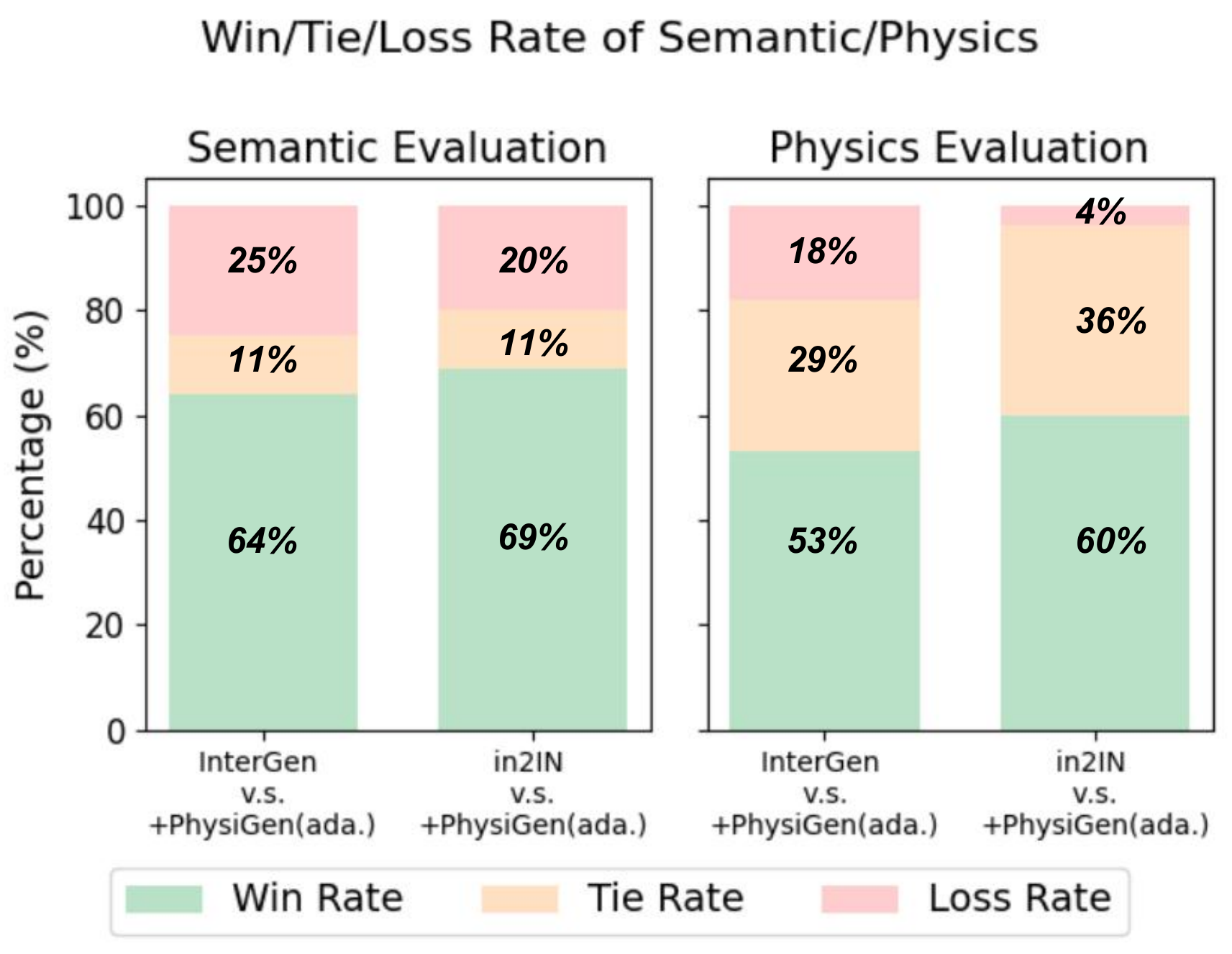}
    \caption{\textbf{User study comparing our method with baselines (InterGen, in2IN).} Participants evaluated the generated video pairs in two ways: how well the motion matched the text (left) and how physically realistic the interaction was (right). The bar chart shows user votes: win rate (green), tie rate (orange), and loss rate (red). The results show that our method was preferred more often and had fewer negative votes in both aspects, meaning it generated better motions in terms of both meaning and physics.}
    \label{fig:user_study}
\end{figure}

In contrast, the SDF loss introduces a significantly higher computational burden. Even with only 128 sampled points, it requires over 24GB of memory—more than 1.5 times that of PhysiGen at its heaviest configuration—and the training time is nearly 7 times longer (0.352 seconds). This becomes drastically worse as the number of points increases, reaching up to 3.7 seconds per batch with 6890 points. The steep resource demand of SDF loss makes it impractical for efficient model training, especially in high-resolution or multi-agent settings.

In summary, our method achieves a much better trade-off between physical realism and computational cost, offering a lightweight and scalable solution suitable for real-world training pipelines.

\subsection{User Study}
To assess the quality of generated motions, we conducted a controlled user study comparing our method with two baselines: InterGen and in2IN.
From the 1,098 test samples in the InterHuman dataset, we randomly selected 100 samples. For each sample, we generated videos using our method and the two baseline models for side-by-side comparison. 

We invited 15 participants to take part in the evaluation. For each video pair, participants compared our result with the baseline. 
They judged the videos from two aspects: whether the motion matched the text (Semantic Evaluation), and whether the interaction showed fewer or no collisions between the characters (Physics Evaluation). If the quality of the two generated videos was hard to judge, they could choose “tie.”

As shown in Fig.~\ref{fig:user_study}, our method is preferred more often than both baselines. For semantic quality, we achieved 64\% and 69\% win rates against InterGen and in2IN, respectively. For physics, the win rates were 53\% and 60\%. These results show that our method produces better motion in both aspects.

\section{Conclusion}

We propose PhysiGen, a plug-and-play, model-agnostic method to reduce physical penetration in human-human motion generation. By modeling body volumes and computing guidance vectors, PhysiGen guides models to produce more physically plausible motions without sacrificing—and sometimes improving—semantic accuracy. Experiments on InterHuman and Inter-X confirm its effectiveness, bridging the gap between realism and meaning in motion generation.

{
    \small
    \bibliographystyle{ieeenat_fullname}
    \bibliography{main}
}

\clearpage
\setcounter{page}{1}
\maketitlesupplementary
\appendix

\section{Appendix}

This supplementary material provides metric definitions, implementation details, additional experiments, and visualizations to support our main paper. We organize the content as follows:

\textbf{In Sec.~\ref{sec:setting}}, we provide more details about the evaluation
metrics and implementations.

\textbf{In Sec.~\ref{sec:vi}}, we present more qualitative results demonstrating the motion generation performance of our method.

\textbf{In Sec.~\ref{sec:limit}}, we discuss the limitations and the potential
research direction based on this work.

\section{More about Experiment Settings}\label{sec:setting}
\subsection{More Details about the Evaluation Metrics}
\noindent \textbf{Frechet Inception Distance (FID):} Features are first extracted from both the generated and real motions, after which the FID is computed by comparing the distributions of these features. As a widely used metric, FID provides a reliable assessment of the overall quality of synthesized motions.

\noindent \textbf{R Precision:} For each generated motion, we construct a description pool containing its ground-truth text description along with Q randomly selected mismatched descriptions from the test set. We then compute the Euclidean distances between the motion features and the text features of all descriptions in the pool and rank them accordingly. The average accuracy is calculated for the top-1, top-2, and top-3 ranks. A retrieval is considered successful if the ground-truth description appears within the top‑k results; otherwise, it is counted as a failure. We set Q = 31 for the Inter‑X dataset and Q = 95 for the InterHuman dataset, following the settings in~\cite{interhuman,related:inter-x}

\noindent \textbf{MM Dist:} The MM distance is defined as the mean Euclidean distance between the motion feature vector of each generated motion and the text feature vector of its corresponding description in the test set.

\noindent \textbf{Diversity:} 
Diversity quantifies the variability of the generated motions. From the entire set of generated motions, two subsets of the same size $S_d$ are randomly sampled. Their respective sets of motion feature vectors $\{\mathbf{v}_1, ..., \mathbf{v}_{S_d} \}$ and $\{\mathbf{v}_1', ..., \mathbf{v}_{S_d}' \}$ are extracted. The diversity of this set of motions is defined as
\begin{equation}
\text{Diversity} = \frac{1}{S_d} \sum_{i=1}^{S_d} \left\| \mathbf{v}_i - \mathbf{v}_i' \right\|_2 .
\label{eq:diversity}
\end{equation}
In our experiments, $S_d = 300$.

\noindent \textbf{MModality:} 
MModality measures the degree of variation among motions generated from the same textual description. For each specific text $c$, two subsets of equal size $S_l$ are randomly sampled from its corresponding generated motions. Their respective sets of motion feature vectors $\{\mathbf{v}_{c,1}, ..., \mathbf{v}_{c,S_l} \}$ and $\{\mathbf{v}_{c,1}', ..., \mathbf{v}_{c,S_l}' \}$ are extracted. The MModality is then computed as:
\begin{equation}
\text{Multimodality} = \frac{1}{C \times S_l} \sum_{c=1}^{C} \sum_{i=1}^{S_l} \left\| \mathbf{v}_{c,i} - \mathbf{v}'_{c,i} \right\|_2 .
\label{eq:multimodality}
\end{equation}
In our experiments, \noindent $S_l = 100$.

\subsection{More Implementation Details}

Following the original implementation, we retain the model structure without modifications.
We use a frozen CLIP-ViT-L/14 model as the text encoder. The motion embedding dimension is set to 1024 for InterGen/in2IN and 512 for TIMotion respectively. During training, the number of diffusion timesteps is set to 1000, and we employ the DDIM~\cite{song2020diffusion} sampling strategy with 50 timesteps and $\eta = 0$. The cosine noise level schedule~\cite{nichol2021cosnoise} and classifier-free guidance~\cite{ho2022classifier} are adopted, with 10\% of random CLIP embeddings set to zero during training and a guidance coefficient of 3.5 during sampling.
All the models are trained using the AdamW~\cite{loshchilov2017adam} optimizer with betas of (0.9, 0.999), a weight decay of $2\times10^{-5}$, a maximum learning rate of $10^{-4}$, and a cosine learning rate schedule with 10 linear warm-up epochs.

We adopt several loss terms from InterGen, including diffusion loss $\mathcal{L}_{\text{simple}}$, foot contact loss $\mathcal{L}_{\text{foot}}$, joint velocity loss $\mathcal{L}_{\text{vel}}$, and bone length loss $\mathcal{L}_{\text{BL}}$.

Finally, the overall loss is defined as:
\begin{equation}
\begin{aligned}
\mathcal{L}_{\text{motion}} &= \mathcal{L}_{\text{simple}} 
+ \lambda_{\text{vel}} \mathcal{L}_{\text{vel}}\\ 
&+ \lambda_{\text{foot}} \mathcal{L}_{\text{foot}} 
+ \lambda_{\text{BL}} \mathcal{L}_{\text{BL}} 
+ \lambda_{\text{coll}} \mathcal{L}_{\text{coll}}.
\end{aligned}
\label{eq:motion_loss}
\end{equation}

\noindent \textbf{InterHuamn dataset:}
To balance the contributions of different loss terms, we set the loss weights as follows for experiments on the InterHuman dataset. For training PhysiGen from scratch, the weights are 
$\lambda_{\text{vel}} = 30$, 
$\lambda_{\text{foot}} = 30$, 
$\lambda_{\text{BL}} = 10$, 
$\lambda_{\text{reg}} = 1$, and 
$\lambda_{\text{coll}} = 0.1$
This model is trained for 2000 epochs with a batch size of 256 on 8 Nvidia 4090 GPUs. 

When fine-tuning a pre-trained model through adaptation, we employ LoRA modules with a rank of 16 and a scaling factor of 32. In this configuration, the loss weights are set 
 $\lambda_{\text{vel}} = 30$, 
$\lambda_{\text{foot}} = 30$, 
$\lambda_{\text{BL}} = 100$, 
$\lambda_{\text{reg}} = 1$, and
$\lambda_{\text{coll}} = 10$, and the model is trained for 500 epochs with a batch size of 256.

\noindent \textbf{Inter-X dataset:}
Existing baselines such as InterGen and TIMotion typically employ a 336-dimensional representation. This vector encodes the 6D rotations for 22 body joints, 3 facial joints, and 30 hand joints, along with the global root joint coordinates and some zero padding.
To apply the proposed physical constraints, our model operates on a 468-dimensional input. This representation includes the 6D rotations of the 22 body and 30 hand joints, in addition to their global coordinates. We decompose this representation into two components for our loss formulation: the part corresponding to the 6D rotations of the 52 joints and the root joint's coordinates is designated as the original representation(ori), while the remaining information is treated as the positional representation(pos).

Consequently, our overall loss function is formulated as a weighted sum of multiple terms, adapted to the richer representation:
\begin{equation}
\begin{aligned}
\mathcal{L}_{\text{motion}} &= \lambda_{\text{ori}}\mathcal{L}_{\text{ori\_simple}} 
+ \lambda_{\text{pos}}\mathcal{L}_{\text{pos\_simple}}  
+ \lambda_{\text{vel}} \mathcal{L}_{\text{vel}} \\
& + \lambda_{\text{foot}} \mathcal{L}_{\text{foot}} 
+ \lambda_{\text{BL}} \mathcal{L}_{\text{BL}} 
+ \lambda_{\text{coll}} \mathcal{L}_{\text{coll}}.
\end{aligned}
\label{eq:motion_loss2}
\end{equation}
During training from scratch, we set specific loss weights for both InterGen-based 
($\lambda_{\text{ori}} = 10$, $\lambda_{\text{coll}} = 10$) and TIMotion-based 
($\lambda_{\text{ori}} = 20$, $\lambda_{\text{coll}} = 50$) models, while others such as 
$\lambda_{\text{vel}} = 30$, $\lambda_{\text{pos}}=1$, $\lambda_{\text{BL}}=10$ and $\lambda_{\text{foot}} = 30$ remained constant. 
These models were trained for 2000 epochs with a batch size of 256 on 8 Nvidia 4090 GPUs.

For adaptation fine-tuning, we set:
 $\lambda_{\text{vel}} = 30$, 
$\lambda_{\text{foot}} = 30$, 
$\lambda_{\text{BL}} = 100$, 
$\lambda_{\text{reg}} = 1$, and
$\lambda_{\text{coll}} = 10$, we employed LoRA modules with a rank of 16 and a scaling factor of 32, training for 500 epochs with the same batch size.

\section{More Qualitative Results}\label{sec:vi}

\begin{figure*}[htbp]
    \centering
    \includegraphics[width=\textwidth]{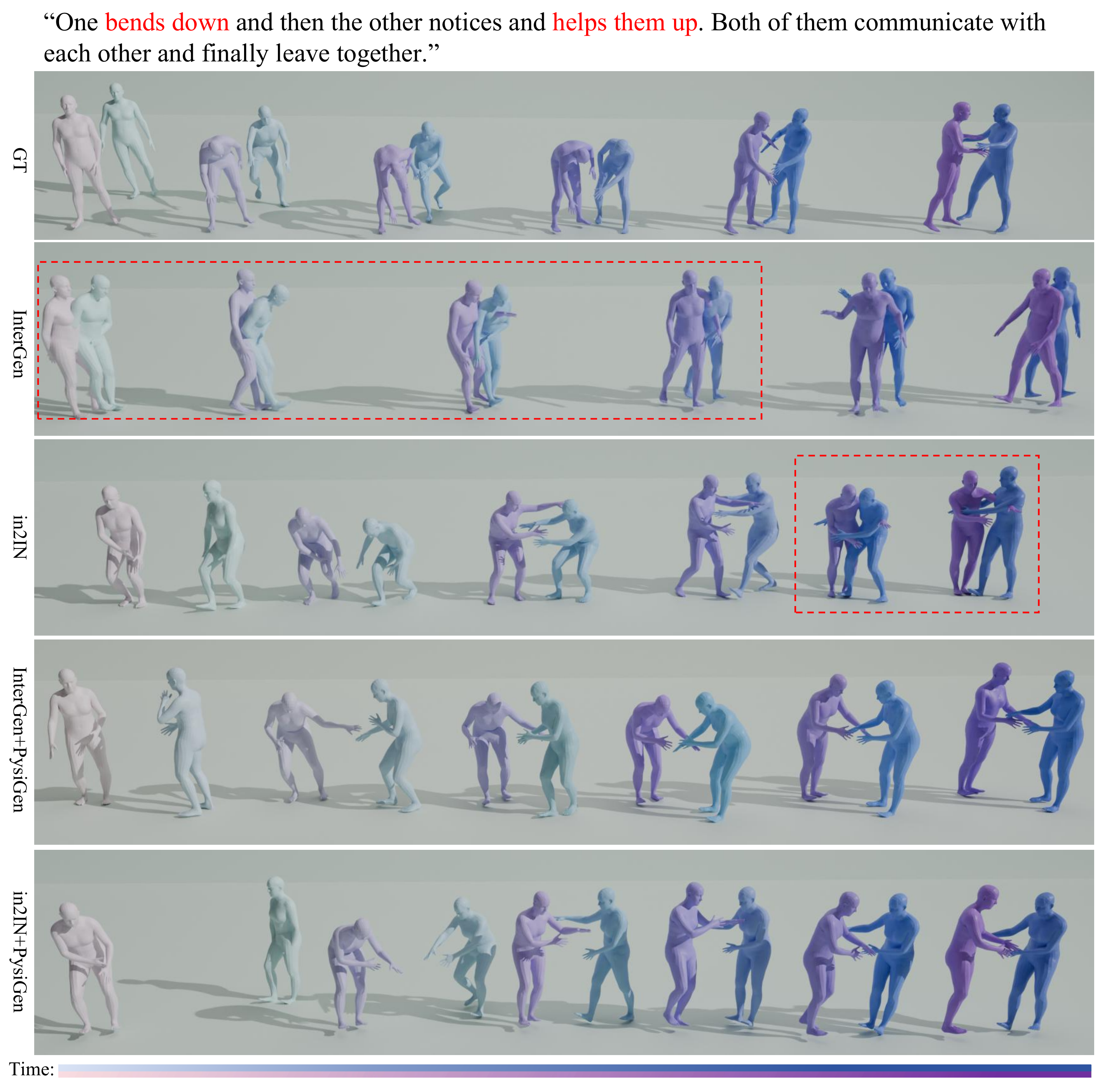}
    \caption{\textbf{Qualitative comparison of generated motions}
Rows show results from GT, InterGen, in2IN, and their versions with PhysiGen. Red dashed boxes highlight severe penetration issues in baseline methods. Models with PhysiGen maintain semantic consistency while significantly improving the physical realism of the generated motions.}
    \label{fig:A2}
\end{figure*}

\begin{figure*}[htbp]
    \centering
    \includegraphics[width=\textwidth]{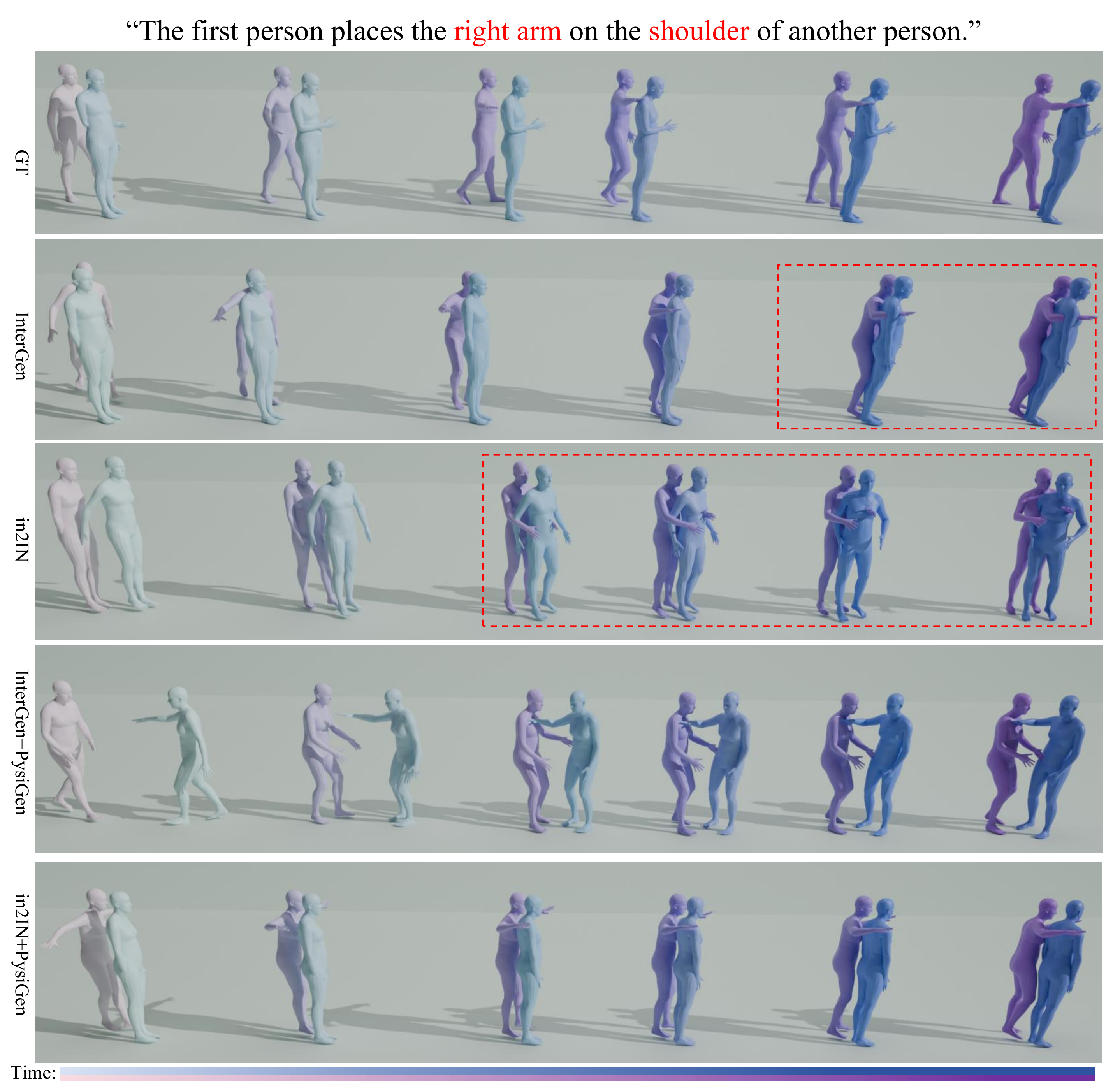}
    \caption{\textbf{Qualitative comparison of generated motions} 
Rows show ground truth (GT), InterGen, in2IN, and their counterparts enhanced with PhysiGen. Red dashed boxes highlight severe collision artifacts in baseline results. 
PhysiGen effectively reduces such collisions while maintaining semantic alignment with the input text.}
    \label{fig:A3}
\end{figure*}

To further demonstrate the effectiveness of our method, we present more qualitative results of motion generation Fig.~\ref{fig:A2},~\ref{fig:A3},~\ref{fig:A4}. These examples compare baseline models with our method, showing better semantic alignment and more realistic physical interactions.

We also provide a demo video package called \textbf{demo.zip}, which contains our generated videos sampled from the InterHuman dataset, which showcase the ability of PhysiGen to reduce penetration and enhance physical realism across various scenarios.

\section{Limitations and Future Work}\label{sec:limit}
\begin{figure*}[htbp]
    \centering
    \includegraphics[width=\textwidth]{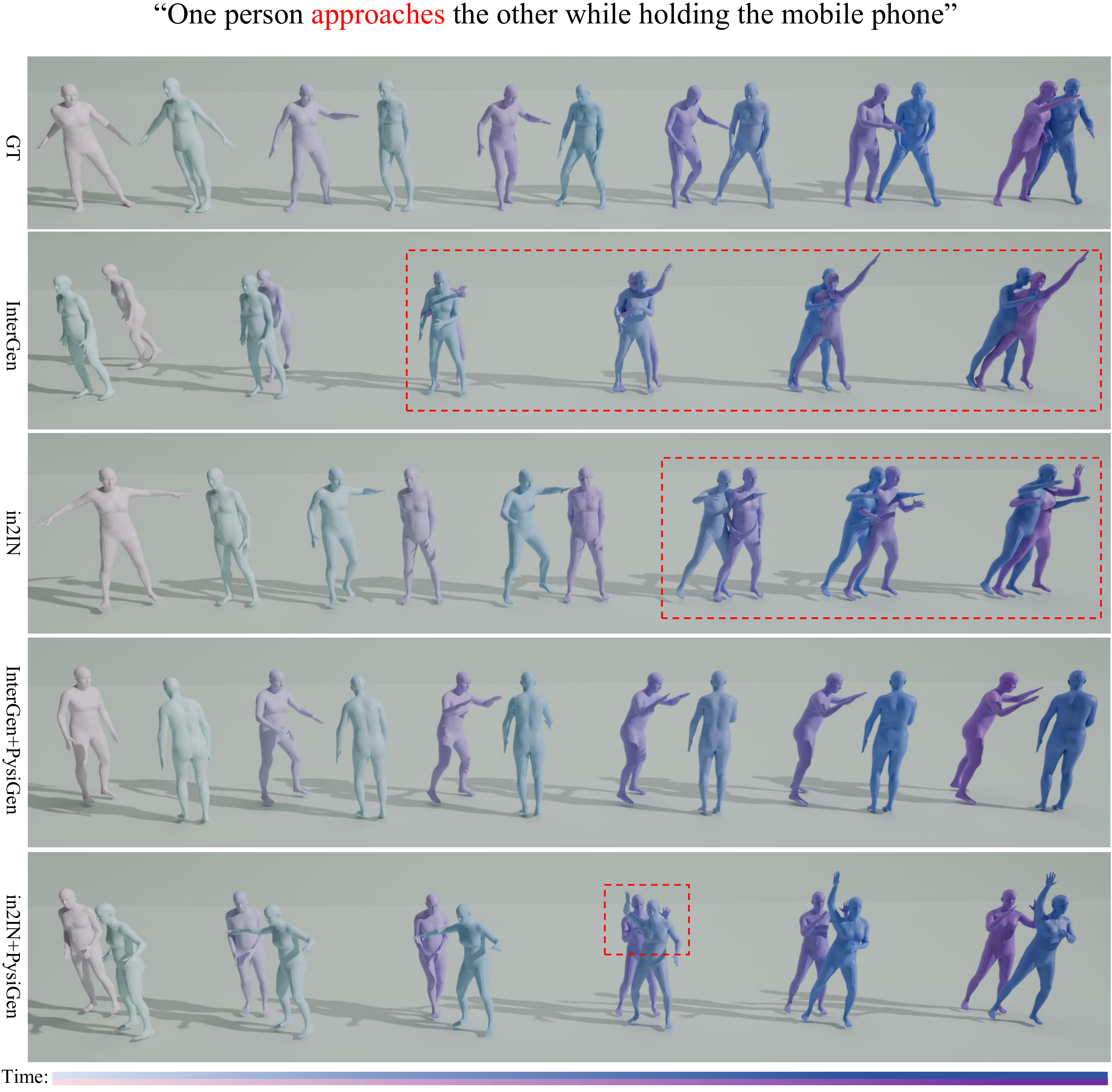}
    \caption{\textbf{Qualitative comparison of generated motions} 
Rows show ground truth (GT), InterGen, in2IN, and their counterparts enhanced with PhysiGen. Red dashed boxes highlight severe collision artifacts in baseline results. 
PhysiGen effectively reduces such collisions while maintaining semantic alignment with the input text.}
    \label{fig:A4}
\end{figure*}

Although our method greatly reduces the probability and severity of collisions, it does not perform fine-grained modeling for fingers. As a result, in some cases (e.g., Fig.~\ref{fig:A4}), the generated hand motions may still penetrate other bodies.
In the future, we plan to improve this work and extend PhysiGen to more complex interaction settings, such as multi-person interactions, crowded scenes, or physical interactions between humans and objects.

\end{document}